\documentclass[lettersize,journal]{IEEEtran}
\usepackage{times}
\usepackage{soul}
\usepackage{url}
\usepackage[hidelinks]{hyperref}
\usepackage[utf8]{inputenc}
\usepackage[small]{caption}
\usepackage{graphicx}
\usepackage{amsmath}
\usepackage{amsthm}
\usepackage{booktabs}
\usepackage{cite}
\usepackage{amsmath,amssymb,amsfonts, optidef}
\usepackage{graphicx}
\usepackage{textcomp}
\usepackage{xcolor}
\usepackage[detect-all]{siunitx}
\usepackage{xspace}
\usepackage{subcaption}
\usepackage{bm}
\usepackage{array}
\usepackage{algorithm,tabularx}
\usepackage{algpseudocode}
\makeatletter
\newcommand{\multilines}[1]{%
	\begin{tabularx}{\dimexpr\linewidth-\ALG@thistlm}[t]{@{}X@{}}
		#1
	\end{tabularx}
}
\makeatother
\usepackage{mathtools}

\usepackage{amsthm}
\usepackage{multirow}
\usepackage{color}
\usepackage{epstopdf}
\usepackage{endnotes}
\usepackage{framed}
\usepackage{cool} 
\usepackage{enumerate} 
\usepackage[subnum]{cases}
\usepackage{multicol}
\usepackage{tablefootnote}
\usepackage{pgf}
\usepackage[detect-all]{siunitx}

\DeclareMathOperator{\dimension}{dim}

\DeclarePairedDelimiterX{\norm}[1]{\lVert}{\rVert}{#1}
\DeclarePairedDelimiterX{\abs}[1]{\lvert}{\rvert}{#1}
\DeclarePairedDelimiterX{\innProd}[1]{\langle}{\rangle}{#1}

\DeclareUnicodeCharacter{FB01}{fi}

\newcommand{\OurAlg}{\ensuremath{\textsf{RBOCPD Pareto klUCB}}\xspace}

\theoremstyle{plain}
\newtheorem{theorem}{Theorem}

\newtheorem{corollary}{Corollary}
\newtheorem{lemma}{Lemma}

\theoremstyle{definition}
\newtheorem{definition}{Definition}

\renewcommand{\baselinestretch}{0.93}
\usepackage{titlesec}
\titlespacing*{\section}{0pt}{0.6\baselineskip}{0.4\baselineskip}
\titlespacing*{\subsection}{0pt}{0.6\baselineskip}{0.4\baselineskip}

\begin{document}
%
\title{Piecewise-Stationary Multi-Objective Multi-Armed Bandit with Application to Joint Communications and Sensing}
\author{Amir Rezaei Balef and Setareh Maghsudi 
\thanks{This work was supported by Grant 01IS20051 and Grant 16KISK035 from the German Federal Ministry of Education and Research (BMBF).}
\thanks{A. Rezaei Balef is with the Department of Computer Science, University of Tübingen, Germany (email: amir.rezaei-balef@uni-tuebingen.de)}
\thanks{S. Maghsudi is with the Department of Computer Science, University of Tübingen, and the Fraunhofer Heinrich-Hertz Institute, Germany
(email: setareh.maghsudi@uni-tuebingen.de)}
}
\maketitle
\begin{abstract}
We study a multi-objective multi-armed bandit problem in a dynamic environment. The problem portrays a decision-maker that sequentially selects an arm from a given set. If selected, each action produces a reward vector, where every element follows a piecewise-stationary Bernoulli distribution. The agent aims at choosing an arm among the Pareto optimal set of arms to minimize its regret. We propose a Pareto generic upper confidence bound (UCB)-based algorithm with change detection to solve this problem. By developing the essential inequalities for multi-dimensional spaces, we establish that our proposal guarantees a regret bound in the order of $\gamma_T\log(T/{\gamma_T})$ when the number of breakpoints $\gamma_T$ is known. Without this assumption, the regret bound of our algorithm is $\gamma_T\log(T)$. Finally, we formulate an energy-efficient waveform design problem in an integrated communication and sensing system as a toy example. Numerical experiments on the toy example and synthetic and real-world datasets demonstrate the efficiency of our policy compared to the current methods.
\end{abstract}
\section{Introduction} 
\label{Sec:Intro}
\IEEEPARstart{M}{}{ulti-Armed Bandit} problem (MAB) is a class of sequential optimization problems. In the most seminal setting, the problem portrays a decision-maker that selects an action from a given set. After selection, it receives some reward generated by the unknown reward process of that arm. Under such uncertainty, at each round, the player may lose some reward (or incur some cost) due to not selecting the best arm instead of the played one. This loss is referred to as regret. The player decides which arm to pull in a sequence of trials to minimize its accumulated regret over the horizon or, alternatively, to maximize its discounted reward \cite{Maghsudi16:MAB}.

In communication systems, many problems involve multiple conflicting objectives. For example, in \cite{kumari2017performance,liu2018mu}, the authors use a multi-objective function to formalize the trade-off between some metrics of a joint communication and sensing (JCAS) system. In the multi-objective setting of the MAB problem, the agent has multiple conflicting objectives; As such, the reward-generating process of each arm returns a vector where each element corresponds to one objective \cite{almasoud2015multi}. Therefore, often the problem has multiple optimal actions and policies when observed from different perspectives, e.g., with various importance over objectives or some other constraints. That aggravates the challenge of efficiently identifying the solution set as one evaluates the policies within a set of solutions. 

The classical approach to deal with the challenge described above is to determine the Pareto front \cite{hayes2021practical} solution set. In \cite{drugan2013designing}, the authors propose three regrets metrics for the multi-objective multi-armed bandit (MoMAB) problems. By introducing the scalarization functions and Pareto search concepts, they develop a decision-making policy to solve the problem. Their results show that Pareto UCB1 performs better than scalarized multi-objective UCB1. As another example, \cite{yahyaa2014knowledge} proposes two methods based on knowledge gradient (KG) that efficiently explores the Pareto optimal arms. The first one is the Pareto order knowledge gradient (Pareto-KG), which uses the Pareto partial order relationship to find Pareto optimal arm set directly in the multi-objective space. The other one, scalarized knowledge gradient (scalarized-KG), defines a set of scalarized functions to generate a variety of elements belonging to the Pareto front set. Those functions convert the multi-dimensions MABs to one-dimension MABs and use the estimated mean and variance for decision-making. The authors also show that their proposal outperforms the Pareto UCB1.

Motivated by several real-world applications \cite{Maghsudi18:PEP}, we study a specific setting of the MoMAB problem, namely, piece-wise stationary MoMAB. In such a setting, the reward-generating processes of arms remain fixed over some intervals and change from one to another. An example is the energy-efficient transmission power adaptation in a JCAS system, where there is a trade-off between energy efficiency (EE) of data information rate (DIR) and probability of detection (PD) and minimum application requirement guarantee. Besides, the channel quality and other relevant network characteristics change over time. 

The non-stationary MAB problem in the single-objective setting has been studied extensively in the past few years. There are two main methods for generalizing the stationary MAB decision-making policies to non-stationary environments. The first method is windowing \cite{garivier2008upper,hayes2021practical}. In this family of methods, one deals with non-stationarity by emphasizing the most recent samples for decision-making. That can be achieved, for example, through discounted weighting, implementing a hard cut-off, and the like.\\
The second method is to deploy a change detector in the decision-making strategy. It identifies the changes in the reward distributions and triggers adaptation. For example, in \cite{liu2018change}, the authors propose a piece-wise stationary MAB framework using a cumulative sum (CUSUM) change detector. In \cite{Maghsudi20:ANB}, the authors use the generalized likelihood ratio (GLR) test to solve the MAB problem with time-varying action availability and apply the results to solve the caching problem.

To our knowledge, only \cite{anquise2021multi} studies the MoMAB problem in a piecewise-stationary setting. The author develops two Pareto UCB policies based on the windowing method described above, namely discounted Pareto UCB (DPUCB) and sliding window Pareto UCB (SWPUCB); nevertheless, the evaluation is only experimental. To our best knowledge, state-of-the-art research has not investigated the piecewise-stationary MoMAB problem using a change-detecting method.

We design a policy for MoMAB problems, namely Pareto generic UCB. We then adapt it to piecewise-stationary environments using a change detector. By developing Cramer-Chernoff’s inequality and Chernoff’s bounds for multi-dimensional spaces, we prove an upper bound for the regret of our proposed method, which shows that Pareto Kullback-Leiber UCB has the best performance. We also establish that our proposed method has a regret bound in the order of $\gamma_T\log(\frac{T}{\gamma_T})$ for a limited number of stationary segments $\gamma_T$ in a piecewise-stationary environment. Finally, we formulate an energy-efficient waveform design problem in a JCAS system as a toy example. The experimental results on the synthetic and real-world datasets demonstrate that our proposed method outperforms the state-of-art policies.
\section{Pareto Generic UCB Policy for MoMAB} 
\label{Sec:problem_formulation}
First, we consider a stationary MoMAB problem with $K \geq 2$ arms and $D$ independent objectives per arm. For the rest of the paper, we use the index $*$ to indicate an optimal parameter. For comparing vectors and measuring the performance, a non-dominated vector and the \textit{Pareto regret}, defined respectively, below.
\begin{definition} (Non-Dominated Vector) 
A vector $\vec{\mu}$ is non-dominated by another one $\vec{\nu}$ ($\vec{\mu} \nsucc \vec{\nu}$)  if and only if $\forall j \in D,\nu^j \geq \mu^j $ and $\exists j \in D,\nu^j >\mu^j$, where $D$ is the dimension of the two vectors. 
\end{definition}
\begin{definition} (Pareto Regret \cite{drugan2013designing}) 
Let $\mathcal{A}^*$ be the \textit{Pareto front set} that includes all arms with a Pareto dominant reward vector. The Pareto regret $\Delta_{i,t}$ measures the distance between the mean vector of the pulled arm $i$ ($\vec{\mu}_i$) at time step $t$ and $\mathcal{A}^*$. Formally,  
\begin{equation}
\Delta_{i,t}=\inf\{\epsilon|\vec{\mu}_{i^*}(t) \nsucc \vec{\mu}_{i}(t) +\epsilon,\forall~i^* \in \mathcal{A}^*\}.
\end{equation}
\end{definition}
Given the definition above, for a policy that selects arm $I_t$ at time step $t$, the cumulative regret follows as
\begin{align}
R(T) = \sum_{t=1}^T \sum_{i=1}^{K} \mathbbm{E} [\mathbbm{1}_{ \{I_t = i \}} \Delta_{i,t}].
\label{eq:expected_regret}
\end{align}
The decision-maker aims at minimizing the cumulative regret.
\subsection{Pareto Generic UCB Policy} 
\label{Sec:CD_GP_UCB}  
Let $t$ be the current time step. We define $f(t) = t {\log^c(t)}, c \geq 1$, and  $\mathcal{D}(\vec{\mu}, \vec{\nu})=\max\limits_{j \in \dimension(\vec{\mu})}d(\mu^j,\nu^j )$, where $d$ is a strong semi-distance KL-dominated function\footnote{See Appendix \ref{sec:app_def} for the definition of a strong semi-distance KL-dominated function.}. At each time step $t$, the agent selects one arm $i$ using \hbox{\textbf{Algorithm \ref*{alg:GP_UCB}}} and receives a reward vector $\vec{r}_i(t)$. The following theorem establishes an upper bound for the regret. 
\begin{theorem} (Regret Bound of Stationary Pareto Generic UCB) 
For \textbf{Algorithm \ref{alg:GP_UCB}}, the upper regret bound up to time $T$ yields
\begin{align}
&R[T] \leq  \sum_{i=1}^{K} (\frac{\log(f(T))}{ \min\limits_{i^* \in \mathcal{A}^* } \mathcal{D}(\vec{\mu}_{i},\vec{\mu}_{i^*})} + O (\log(\log(T)))  \Delta_{i,T}.
\end{align} 
\label{theorem:regret}
\end{theorem}
\begin{algorithm}[t!]
\caption{Pareto Generic UCB Policy}
\label{alg:GP_UCB}
\begin{algorithmic}[1]
\If {$t\leq K$}
\State Randomly pulls an arm that has not been chosen so far.
\Else
\For {all arms $i \in K$}
\State \parbox[t]{0.8\linewidth}{$N_i(t)=\sum_{m=1}^{t} \mathbbm{1}_{\{I_m=i\}}$ \Comment{$N_i$ is the number of times arm $i$ has been selected at step $t$.}}
\State \parbox[t]{0.8\linewidth}{Calculate the empirical mean reward of $\vec{\mu}_i$ as \hbox{$\vec{\mu}_i(t)=\frac{\sum\limits_{m=1}^{t}\vec{r}_i(m)\mathbbm{1}_{ \{I_m = i \}}}{N_i(t)}$}.}
\State Define $\delta_i(t)=\frac{\log(f(t))}{N_i(t)}$.
\State \parbox[t]{0.8\linewidth}{Calculate the UCB vector of arm $i$ as \hbox{$\vec{U}_i=\sup \{ \vec{u}:\mathcal{D}(\vec{\mu}_i( t),\vec{u}) \leq \delta_i(t)\}$}.}
\EndFor
\State $\mathcal{A}^*=\{i|\forall~l \in K,\vec{U}_l \nsucc \vec{U}_i\}$
\State Select an arm $i$ uniformly at random from $\mathcal{A}^*$.
\EndIf 
\end{algorithmic}
\end{algorithm}
\section{Piecewise-stationary MoMABs} 
\label{Sec:CD}
We use a change-point detection framework to adapt the stationary MoMAB algorithm to a piecewise-stationary environment.
\textbf{Algorithm \ref*{alg:CD_bandits}} controls \textbf{Algorithm \ref*{alg:GP_UCB}} as follows: if it detects a change, it restarts \textbf{Algorithm \ref*{alg:GP_UCB}} by setting $t$ to $1$. For detecting changes in the distribution of rewards a change detection algorithm is required. We use the \textit{Restarted Bayesian Online Change-point Detector} (RBOCD) \cite{alami2020restarted} method. Let $\delta<<1$. Besides, $\varDelta$ represents the minimum detectable gap. Then, the probability of false alarm and the upper bound of detection delay are respectively given by 
\begin{gather}
\mathbbm{P}\{RBOCD = 1\} \leq \delta, \quad
D_{\varDelta } \leq \frac{o(\log \frac{1}{\delta})}{2\varDelta^2}.\notag
\end{gather}
Thus, $\delta$ controls the trade-off between false alarm probability and detection delay. The upper bounds for expected values of false alarm and detection delay respectively follow as
\begin{gather}
\mathbbm{E}[F] \leq \delta T,\quad \mathbbm{E}[D_{\varDelta }] \leq \frac{o(\log \frac{1}{\delta})}{2\varDelta^2}.
\label{eq:RBOCD_bounds}
\end{gather}
%
\begin{theorem} (Regret Bound of Piecewise-stationary Pareto Generic UCB) 
If $R(T) = O(\log(T))$ (as for Algorithm \ref*{alg:GP_UCB}), the total regret bound of Algorithm \ref*{alg:CD_bandits} yields
\begin{align}
R_{tot}(T)=O\left((\gamma_T+\mathbbm{E}[F])\log(\frac{T}{\gamma_T+\mathbbm{E}[F]}) +\gamma_T \mathbbm{E}[D_{\varDelta }]\right). 
\end{align}
\label{theorem:regret_total}
\end{theorem}
\begin{algorithm}[t!]
\caption{Piecewise-stationary MoMABs}
\label{alg:CD_bandits}
\begin{algorithmic}[1]
\State $\tau \gets 1; t \gets 1$
\While {$\tau \leq T $}
\State Run the policy (Algorithm \ref*{alg:GP_UCB})
\State  $ \tau \gets \tau+1; t \gets t+1$   
\If {$CD(\vec{r}_i(t),i)$=True}
\State Restart the policy (Algorithm \ref*{alg:GP_UCB}) by $t \gets 1$
\EndIf
\EndWhile
\end{algorithmic}
\end{algorithm}
\begin{corollary} (ROCBD Pareto KL-UCB) 
In ROCBD, by choosing the Kullback-Leiber distance function and $\delta = \frac{1}{T}$, we have $R(T)=O(\gamma_T\log(T))$. Besides, if the number of breakpoints is a priori known,\footnote{See \cite{liu2018change} (Remark 2) for justification that such assumption is reasonable.} $\delta = \frac{\gamma_T}{T}$ results in $R(T) = O(\gamma_T\log(\frac{T}{\gamma_T}))$.
\end{corollary}
\section{Numerical Experiments}
\label{Sec:experiments}
We evaluate the regret performance of our proposal (\OurAlg) compared to the state-of-the-art methods using the toy example described below, and also with real and synthetic datasets. We select $f(t)=t$. As benchmarks, we use several methods, including DPUCB and SWPUCB \cite{anquise2021multi}, as described in \textbf{Section \ref{Sec:Intro}}. We use grid search to determine the hyper-parameters of DPUCB and SWUCB, i.e., the discount factor and the window length.

\begin{figure}
\centering
\begin{subfigure}{.5\linewidth}
\centering
\resizebox{1\linewidth}{!}{\includegraphics{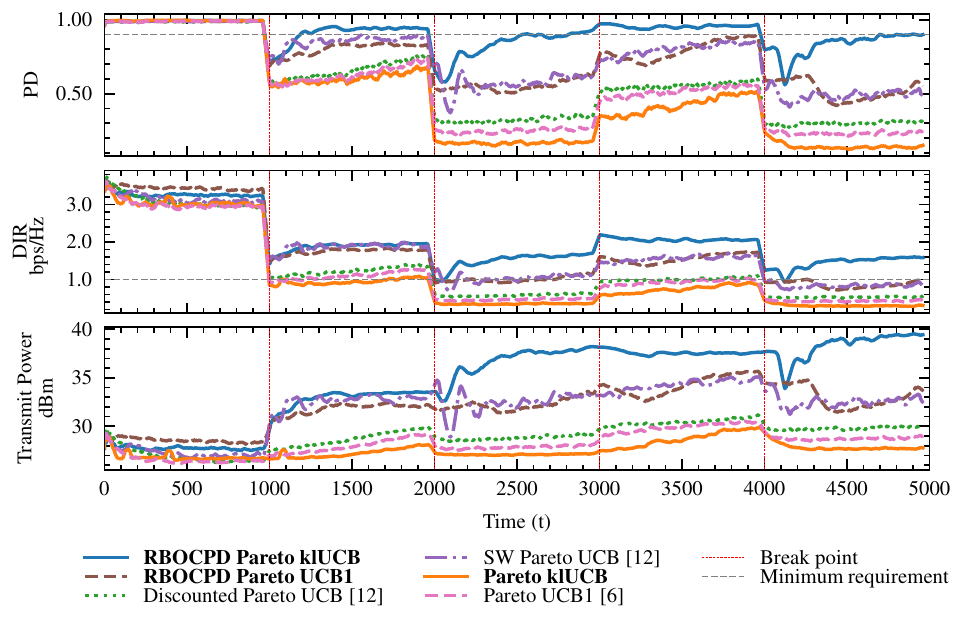}}
\caption{Performances}
\label{fig:JCAS_Performances}
\end{subfigure}%
\begin{subfigure}{.5\linewidth}
\centering
\resizebox{1\linewidth}{!}{\includegraphics{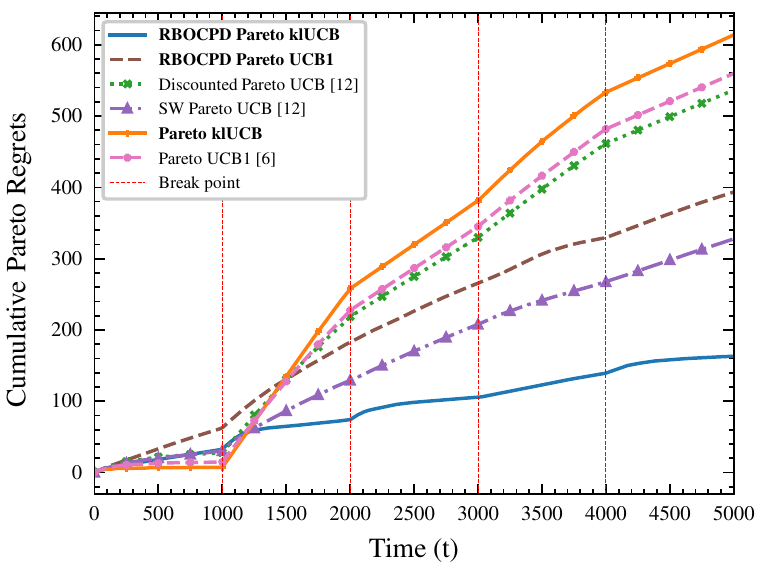}}
\caption{Regret} 
\label{fig:JCAS_regrets}	
\end{subfigure}
\caption{ Comparison of regret and performances for the toy example. After some iterations, our proposed algorithm converges to the optimal solutions that satisfy the minimum desired requirements. That means that our algorithm is applicable in much practical communication as an online optimizer.}
\label{fig:JCAS}
\end{figure}

\begin{figure}
\centering
\begin{subfigure}{.5\linewidth}
\centering
\resizebox{1\linewidth}{!}{\includegraphics{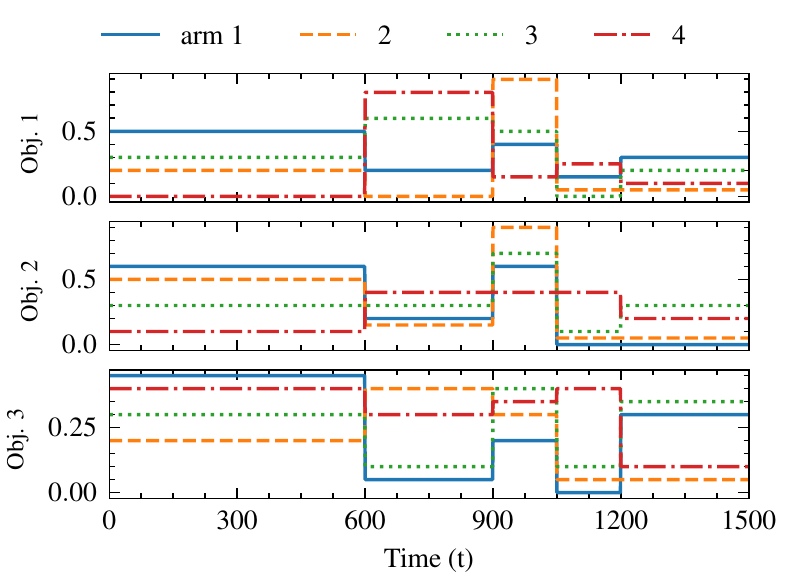}}
\caption{Mean Reward}
\label{fig:synthetic_means}
\end{subfigure}%
\begin{subfigure}{.5\linewidth}
\centering
\resizebox{1\linewidth}{!}{\includegraphics{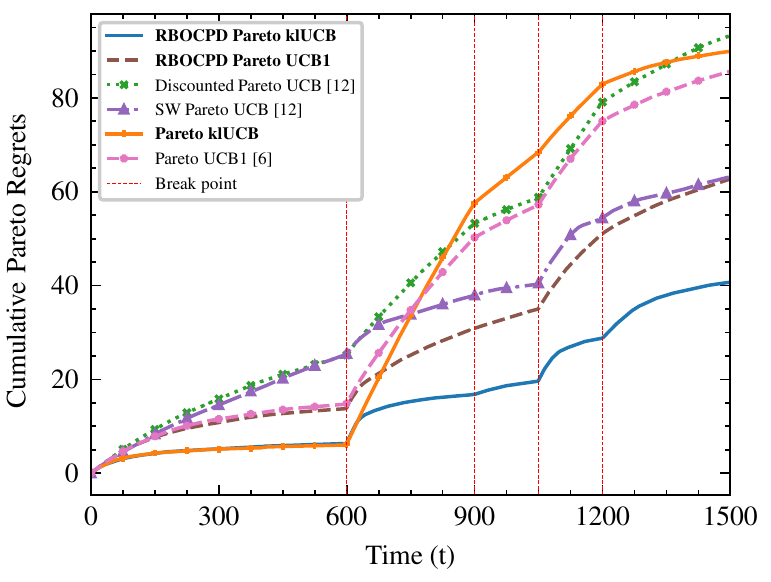}}
\caption{Regret} 
\label{fig:synthetic_regrets}	
\end{subfigure}
\caption{Comparison of regret performance using a synthetic dataset with $K=4$ arms, $\gamma_T=4$ breakpoints and $D=3$ objectives.}
\label{fig:synthetic}
\end{figure}

\begin{figure}
\centering
\begin{subfigure}{.5\linewidth}
\centering
\resizebox{1\linewidth}{!}{\includegraphics{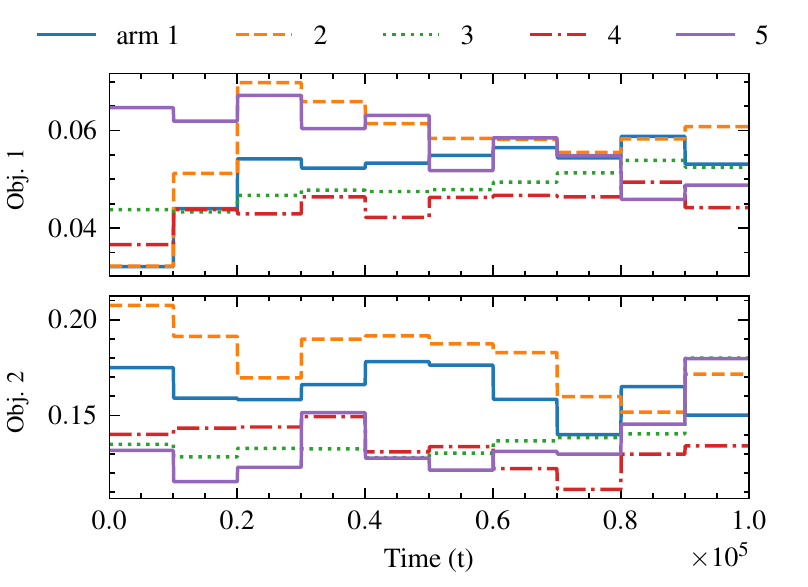}}
\caption{Mean Reward}
\label{fig:real_means}
\end{subfigure}%
\begin{subfigure}{.5\linewidth}
\centering
\resizebox{1\linewidth}{!}{\includegraphics{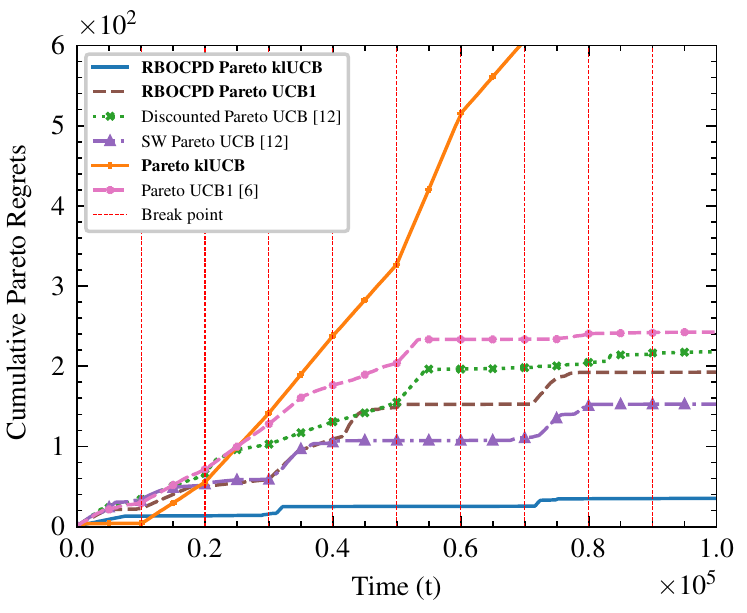}}
\caption{Regret} 
\label{fig:real_regrets}	
\end{subfigure}
\caption{ Comparison of regret performance using the Yahoo! Webscope dataset with $K=10$ arms, $\gamma_T= 9$ breakpoints, and $D=2$ objectives. The mean rewards corresponding to 5 out of 10 articles appear in the figure.}
\label{fig:real}
\end{figure}

\textbf{Toy Example:} We consider a JCAS system that transmits a single waveform to a radar target and communication user. We aim to find the best EE of two performance metrics, the PD and the DIR, in the JCAS system using MoMAB algorithms. Assume $P_t$ is the transmission power whose value belongs to a set with $K$ elements $\{P_{\min}+\frac{k}{K-1} (P_{\max}-P_{\min})|\forall k \in \{ 0, 1, ..., K-1 \}\}$. For a cellular communication system, DIR is given by
\begin{gather}
{\mathrm{DIR}}=\log_{2}\left(1+ P_t G_c / N_0\right),
\end{gather}
where $G_c$ is the communication channel gain, and $N_0$ is the power spectral density (PSD) of the AWGN noise.\\
Among several performance metrics, we use the probability of false alarm (PFA) and PD, respectively, defined as \cite{chalise2017performance} 
\begin{gather}
{\mathrm{PD}}\approx Q_{1}\left(\sqrt{2 P_t G_r / N_0 },\sqrt{2\lambda}\right),
\end{gather}
where $\lambda$ is the control parameter for the false alarm rate ${\mathrm{PFA}}\approx\mathrm{e}^{-\lambda}$. The radar channel gain at the target is $Gr$.\\
The power consumption model follows as \cite{bjornson2018energy} 
\begin{gather}
P_{tot} = \omega + P_t + \nu B + \eta B (\mathrm{DIR}),
\end{gather}
where $\nu$ and $\eta$ are proportional to the ADC/DAC and encoder/decoder energy consumption. They are also related to bandwidth $B$, the required circuit power for link operation $\omega$, and the DIR. The optimization problem yields
\begin{maxi!}
  {P_t}{ ( \frac{{\mathrm{DIR}}}{P_{tot}}, \frac{{\mathrm{PD}}}{P_{tot}}) \label{eq:optProb}}{}{}
  \addConstraint{{\mathrm{DIR}}}{ \geq  \mathrm{DIR}_{\min} \label{eq:constraint1}}
  \addConstraint{{\mathrm{PD}}}{\geq  \mathrm{PD}_{\min}  
  \label{eq:constraint2}}
\end{maxi!}
We set the parameter as follows: The carrier frequency and bandwidth are 60 \si[]{\giga\hertz} and 2.6 \si[]{\giga\hertz}, respectively. $n=2$ is the path-loss exponent \cite{kumari2017performance}. Channel fading follows a Rayleigh distribution with a variance of $15$ \si{\dB}. We run experiments for $\omega = 2$ \si[]{\watt}, $N_0 = -174$ \si[per-mode=symbol]{{\dB\milli}\per\hertz}, $\nu = 10^{-14}$ \si[]{\joule}, and $\eta = 10^{-15}$ \si[per-mode=symbol]{\joule\per bit} \cite{bjornson2018energy}. The transmission power changes between $20$ to $40$ \si[]{\dB\milli\relax} in $2.5$ \si[]{\dB\milli\relax} steps; That is, the MAB model involves $K=9$ arms. Moreover, we select $\lambda = 5$ so the probability of false alarm is $\mathrm{PFA} \approx e^{-5} < 0.01$ and the minimum requirements are determined by $\mathrm{DIR}_{\min}= 1$ and $\mathrm{PD}_{\min} = 0.9 $. To model non-stationarity, we assume $4$ breakpoints in time horizon $T=5000$ and we start the target distance with $l =10$. After each breakpoint, we subsequently assign $40, 80, 60$, and $100$ to $l$. \textbf{Figure \ref{fig:JCAS_Performances}} and \textbf{Figure \ref{fig:JCAS_regrets}} show the mean of the performance metrics and the cumulative Pareto regrets over 100 runs, respectively.\\
\textbf{Synthetic Dataset:} 
The synthetic dataset has three objectives and four arms, each with a Bernoulli distribution. \hbox{\textbf{Figure \ref{fig:synthetic_means}}} shows the mean of the Bernoulli distribution of arms. Furthermore, there are four breakpoints. We select $T=1500$. We report the average of cumulative Pareto regrets over 100 runs in \hbox{\textbf{Figure \ref{fig:synthetic_regrets}}}.\\
\textbf{Real Dataset:} 
We use the Yahoo! Webscope dataset (R6A), which consists of over 45 million visits to the Yahoo! Today module for ten days. We consider the data collected on the first day, which includes around 4.5 million samples. In \cite{tekin2018multi}, the authors used the click-through rate and the average payment as objectives. The average payment is based on the pay-per-view model, implying that the advertiser makes a unit payment to the publisher for each displayed ad. We use a similar approach to have bi-objectives rewards but with a slight modification to create a Bernoulli distribution. In the first step, we choose 10 articles that have the highest number of records in the dataset which means they have the maximum durability and overlapped in time the most, after that for having a non-stationary Bernoulli reward for the number of views we give reward 1 to the most repeated article in segments with a length of 5 in the whole dataset. We show the mean reward of 5 randomly-selected articles in \textbf{Figure \ref*{fig:real_means}}. The piecewise-stationary length is 10000 and the time horizon is $T=100000$.\\
\section{Conclusions}
In this letter, we proposed an algorithm to solve a piece-wise stationary multi-objective MAB problem. We analyze our proposal theoretically. Besides, experimental results show the superior performance of our proposed methods over the state-of-the-art algorithms. We then provide an example of a communication problem solvable by our scheme, namely the transmission power adaptation problem in a JCAS system. Another example is optimizing many-to-many communication in cognitive radio networks regarding delay performance, link occupancy,  and QoS guarantee. For future research, one could investigate the proposed scheme under optimizing beamforming design for JCAS or phase shifts for re-configurable intelligent surfaces (RIS). Moreover, at the algorithm level, one can develop it for off-policy learning scenarios.
\appendix
\subsection{Definitions}
\label{sec:app_def}
\begin{definition} (Strong Semi-Distance KL-dominated Function \cite{liu2018ucboost}) 
Let $d_{kl}$ be a Kullback–Leibler distance function.  A function $d$ is a strong semi-distance KL-dominated function if for $p,q \in \Theta$, we have
\begin{itemize}
\item $0 \leq d(p,q) \leq d_{kl}(p,q) $
\item $d(p,q)=0$, if and only if $p=q$
\item $d(p,q) \geq d(p^\prime,q), \forall p \leq p^\prime \leq q $ 
\item $d(p,q) \leq d(p,q^\prime), \forall p \leq q \leq q^\prime $
\end{itemize}
\label{def:kl_dominated}
\end{definition}
\subsection{Inequalities for \texorpdfstring{$D$}-dimensional Spaces}
To bound the regret of our proposed algorithm, we first generalize some known bounds to the $D$-dimensional spaces.
\begin{lemma} (Extended Cramer-Chernoff's Inequality) 
Let $\vec{X}: S \rightarrow R^D$ be a sequence of i.i.d non-negative random numbers. Consider $\vec{A} \in R^D$. Then $\forall~A^j>0$ and $r>0$, we have
\begin{gather}
\mathbbm{P} \{ \vec{X} \nprec \vec{A} \} \leq  \min_{j \in D} \mathbbm{P} \{ {X^j} \geq {A^j} \}  \leq \min_{j \in D}\frac{\mathbbm{E}[{rX^j} ]}{{rA^j}}
\end{gather}
\label{lemma:Ineq_Cramer_Chernoff}
\end{lemma}
\begin{lemma} (Extended Chernoff's Bound) 
Let $\vec{X}: S \rightarrow R^D$ be a sequence of i.i.d Bernoulli random variables. Besides, $\vec{\mu} = \mathbbm{E}[\vec{X}]$, $\vec{\hat{\mu}} = \frac{1}{n}\sum_{i=1}^{n}\vec{X_i}$, $\vec{\epsilon} \in R^D$ with $\epsilon^j>0$. We have
\begin{gather}
\mathbbm{P} \{\vec{\hat{\mu}} \nprec \vec{\mu}+\vec{\epsilon}\} \leq \exp(-n\mathcal{D}(\vec{\mu},\vec{\mu}+\vec{\epsilon}))
\label{eq:Chernoff_Bounds_a}
\end{gather}
\begin{gather}
\mathbbm{P} \{\vec{\hat{\mu}} \nsucc \vec{\mu}-\vec{\epsilon}\} \leq  \exp(-n\mathcal{D}(\vec{\mu}-\vec{\epsilon},\vec{\mu})) 
\label{eq:Chernoff_Bounds_b}
\end{gather}
\begin{gather}
\mathbbm{P}\{\mathcal{D}(\vec{\hat{\mu}},\vec{\mu}) \geq \epsilon , \vec{\mu} \nprec \vec{\hat{\mu}}\} \leq e^{-n \epsilon}
\label{eq:Chernoff_Bounds_c}
\end{gather}
\begin{gather}
\mathbbm{P} \{\mathcal{D}(\vec{\mu},\vec{\hat{\mu}}) \geq \epsilon , \vec{\hat{\mu}} \nprec \vec{\mu}\} \leq e^{-n\epsilon}
\label{eq:Chernoff_Bounds_d}
\end{gather}
\begin{gather}
\mathbbm{P} \{\vec{\mu} \nprec \vec{U}(\epsilon)\} \leq e^{-n \epsilon}
\label{eq:Chernoff_Bounds_e}
\end{gather}
where $\vec{U}(\epsilon)=\sup\{\vec{u}:\mathcal{D}(\vec{\hat{\mu }}, \vec{u}) \leq \epsilon\}$.
\label{lemma:Chernoff_Bounds}
\end{lemma}
\begin{IEEEproof}
First, we prove (\ref{eq:Chernoff_Bounds_a}). The proof of (\ref{eq:Chernoff_Bounds_b}) is similar.
\begin{gather}
\mathbbm{P} \{\vec{\hat{\mu}} \nprec \vec{\mu}+\vec{\epsilon}\} 
=\mathbbm{P}\{e^{\lambda \sum_{i=1}^{n}\vec{X_i}} \nprec e^{n \lambda( \vec{\mu} + \vec{\epsilon})}\} \notag \\
\leq \min_{j \in D} (e^{- n\lambda (\mu^j+\epsilon^j)}\mathbbm{E}[ \prod_{i=1}^{n} e^{\lambda X_i^j}]). 
\end{gather}
From Lemma 9 of \cite{garivier2011kl}, for $x \in [0,1]$, we have $\mathbbm{E}[e^{\lambda x}] \leq  1-\mu+\mu e^\lambda$. Hence,
\begin{gather}
\mathbbm{P} \{\vec{\hat{\mu}} \nprec \vec{\mu}+\vec{\epsilon} \} \leq \min_{j \in D} (e^{-n \lambda(\mu^j +\epsilon^j )}(1- \mu^j + \mu^j e^\lambda)^n ) \notag \\ 
= \min_{j \in D}(e^{n \phi^j(\lambda)})=\exp(n \min_{j \in D}\phi^j(\lambda)),
\label{neq:Chernoff_Bounds}
\end{gather}
where $\phi^j(\lambda)=\log[e^{-\lambda(\mu^j+\epsilon^j)}(1-\mu^j+\mu^j e^\lambda)]$. To find the minimum of $\phi^j(\lambda)$, we set $\frac{d\phi^j(\lambda)}{d\lambda}=0$, which yields
\begin{gather}
\lambda_{\phi^j_{min}}= \log \frac{(\mu^j+\epsilon^j)(1-\mu^j)}{\mu^j(1-\mu^j-\epsilon^j)}.
\label{eq:lambda_min}
\end{gather}
Finally, by combining (\ref*{eq:lambda_min}) and (\ref*{neq:Chernoff_Bounds}), we conclude that
\begin{gather}
\mathbbm{P} \{\vec{\hat{\mu}} \nprec \vec{\mu}+\vec{\epsilon}\} \leq \exp(- n  \max_{j \in D}d_{kl}(\vec{\mu},\vec{\mu}+\vec{\epsilon}))\notag \\
\leq \exp(-n\mathcal{D}(\vec{\mu},\vec{\mu}+\vec{\epsilon})). \notag 
\end{gather}    
Next, we prove (\ref*{eq:Chernoff_Bounds_c}). The proof of (\ref*{eq:Chernoff_Bounds_d}) is similar. \\
Suppose $x$ is a solution to $\mathcal{D}(\vec{\mu}-x ,\vec{\mu})=\epsilon$. Then,
\begin{gather}
\{d(\vec{\hat{\mu}},\vec{\mu})\geq \epsilon,\vec{\mu} \nprec \vec{\hat{\mu}}\}=\{\mathcal{D}(\vec{\hat{\mu}},\vec{\mu}) \geq \mathcal{D}( \vec{\mu} -x,\vec{\mu} ),\vec{\mu}\nprec\vec{\hat{\mu}} \}\notag \\ 
= \{\vec{\mu}-x \nprec \vec{\hat{\mu}},\vec{\mu}\nprec\vec{\hat{\mu}}\} = \{\vec{\mu}-x \nprec \vec{\hat{\mu}}\}. \notag
\end{gather} 
From (\ref*{eq:Chernoff_Bounds_b}), we have
\begin{gather}
\mathbbm{P} \{\vec{\mu}-x \nprec \vec{\hat{\mu}}\} \leq  \exp(- n\mathcal{D}(\vec{\mu} -x,\vec{\mu} )) \leq  \exp(- n\epsilon). \notag
\end{gather} 
Finally, to prove (\ref*{eq:Chernoff_Bounds_e}), we only need to show that the condition of (\ref*{eq:Chernoff_Bounds_e}) is a subset of the condition of (\ref*{eq:Chernoff_Bounds_c}).
\begin{gather}
\{\vec{\mu} \nprec \vec{U}(\epsilon)\}=\{\vec{\mu} \nprec  \vec{U}(\epsilon),\vec{\mu}\nprec \vec{\hat{\mu}}\} \notag\\ 
=\{\mathcal{D}(\vec{\hat{\mu}},\vec{\mu}) \geq \mathcal{D}(\vec{\hat{\mu}} , \vec{u}),\vec{\mu} \nprec \vec{\hat{\mu}}\} \notag\\
\leq \{\mathcal{D}( \vec{\hat{\mu}},\vec{\mu}) \geq \epsilon,\vec{\mu} \nprec \vec{\hat{\mu}}\}.  \notag
\end{gather}
\end{IEEEproof}
\subsection{Proof of Theorem \ref{theorem:regret}}
In the first step, we upper bound the number of times a sub-optimal arm $i \neq i^* \in \mathcal{A}^*$ is played.
\begin{align}
 &\mathbbm{E}[\tilde N_i(T)] \leq L_i+\sum_{t= L_i}^{T}\mathbbm{1}_{\{I_t=i,N_i(t) \geq L_i\}}
\label{eq:kl_N_i_A_i}
\end{align}
Where $L_i>0$. Besides, by assumption, the sub-optimal arm $i$ has been played already $L_i$ times. The second
term on the right-hand side of (\ref{eq:kl_N_i_A_i}) is the number of times the algorithm chooses the sub-optimal arm $i$ for $N_i(t) \geq L_i$. Based on \hbox{\textbf{Algorithm \ref*{alg:GP_UCB}}} (Lines 10 and 11), we can rewrite- and decompose that term as
\begin{align}
&\sum_{t= L_i}^{T}\mathbbm{1}_{\{I_t=i,N_i(t) \geq L_i\}} =  \sum_{t=L_i}^{T} \frac{\sum\limits_{i^* \in \mathcal{A}^*} \mathbbm{1}_{\{\vec{U}_{i^*}(t) \nsucc \vec{U}_i(t),N_i(t) \geq L_i\}}}{|\mathcal{A}^*|} \notag \\
&\leq \sum_{t= L_i}^{T} \frac{1}{|\mathcal{A}^*|} \sum_{i^* \in \mathcal{A}^* }  \mathbbm{1}_{\{\vec{\mu}_{i^*} \nprec \vec{U}_i(t),N_i(t) \geq L_i\} } \notag \\
&+\sum_{t= L_i}^{T} \frac{1}{|\mathcal{A}^*|} \sum_{i^* \in \mathcal{A}^* } \mathbbm{1}_{\{ \vec{\mu}_{i^*} \prec \vec{U}_i(t),\vec{U}_{i^*}(t) \nsucc \vec{U}_i(t) ,N_i(t) \geq L_i\}}\notag 
\label{eq:kl_N_i_A_i_dec}
\end{align}
For the condition of the first term on the right-hand-side of (\ref{eq:kl_N_i_A_i_dec}), from (\ref*{eq:Chernoff_Bounds_e}) of Lemma \ref{lemma:Chernoff_Bounds}, we have
\begin{align}
\mathbbm{P} \{\vec{\mu}_{i^*} \nprec \vec{U}_i(t)\} &\leq e^{-N_i(t) \frac{\log(f(t))}{N_i(t)}} = \frac{1}{f(t)}
\end{align} 
For the condition of the second term on the right-hand-side of (\ref{eq:kl_N_i_A_i_dec}), we have
\begin{align*} 
&\mathbbm{P}\{\vec{\mu}_{i^*} \prec \vec{U}_i(t),\vec{U}_{i^*}(t) \nsucc \vec{U}_i(t) , N_i(t) \geq L_i\} \notag\\ 
&\leq P \{\vec{\mu}_{i^*} \prec \vec{u},\mathcal{D}(\vec{\mu}_{i}(t),\vec{u}) \leq \delta_i(t),N_i(t) \geq L_i\} \notag \\ 
\end{align*}
From $\vec{\mu}_{i}(t) \nsucc \vec{\mu}_{i^*} \prec \vec{u}$ and   $N_i(t) \geq L_i$, it follows that $\mathcal{D}(\vec{\mu}_{i}(t),\vec{\mu}_{i^*}) \leq \mathcal{D}(\vec{\mu}_{i}(t),\vec{u})$ and $\delta_i(t) \leq \frac{\log(f(t))}{L_i}$, respectively. Let $\vec{\epsilon}$ be a positive arbitrary vector such that  $\vec{\epsilon}  \nsucc \vec{\mu}_{i^*}- \vec{\mu}_{i}$. Then, by choosing $L_i =\frac{\log(f(T))}{ \min\limits_{i^* \in \mathcal{A}^* } \mathcal{D}(\vec{\mu}_{i}+\vec{\epsilon},\vec{\mu}_{i^*})}$, we have
\begin{align}
& P \{\vec{\mu}_{i^*} \prec \vec{u},\mathcal{D}(\vec{\mu}_{i}(t),\vec{u}) \leq \delta_i(t),N_i(t) \geq L_i\} \notag \\ 
&\leq   P \{ \mathcal{D}(\vec{\mu}_{i}(t),\vec{\mu}_{i^*}) \leq \frac{\log(f(T))}{L_i}  \}  \notag \\ 
&\leq  P \{ \mathcal{D}(\vec{\mu}_{i}(t),\vec{\mu}_{i^*}) \leq \min_{i^* \in \mathcal{A}^*} \mathcal{D}(\vec{\mu}_{i} +  \vec{\epsilon} ,\vec{\mu}_{i^*}) \} \notag \\
&\leq P \{ \mathcal{D}(\vec{\mu}_{i}(t),\vec{\mu}_{i^*}) \leq \mathcal{D}(\vec{\mu}_{i} + \vec{\epsilon} ,\vec{\mu}_{i^*}) \} \notag \\ 
&\leq P \{ \vec{\mu}_{i}(t) \nprec \vec{\mu}_{i} + \vec{\epsilon} \}\leq e^{-t \mathcal{D}( \vec{\mu}_{i},\vec{\mu}_{i} + \vec{\epsilon} ) }
\end{align} 
By putting everything together, the regret bound for \hbox{$f(t) \geq t \log t$} is:
\begin{align}
&\mathbbm{E}[\tilde N_i(T)]\leq L_i+\sum_{s=L_i}^{T}(\frac{1}{f(s)}+e^{-s\mathcal{D}( \vec{\mu}_{i},\vec{\mu}_{i}+\vec{\epsilon})})\notag \\ 
&\leq L_i +\log( \frac{\log(T)}{\log(L_i)}) + \frac{e^{-L_i}-e^{-T}}{e^{\mathcal{D}( \vec{\mu}_{i},\vec{\mu}_{i}+\vec{\epsilon}))}-1 }
\end{align}
For $\vec{\epsilon} \rightarrow 0 $,  we  have:
\begin{align}
&R[T] \leq  \sum_{i=1}^{K} (\frac{\log(f(T))}{ \min\limits_{i^* \in \mathcal{A}^* } \mathcal{D}(\vec{\mu}_{i} ,\vec{\mu}_{i^*})} + O (\log(\log(T))) \Delta_{i,T}
\end{align}
%
\subsection{Proof of Theorem \ref{theorem:regret_total}}
Let $F_i$ and $\gamma_T$ be the number of false alarms and breakpoints up to time $T$, respectively. Besides, $D_i^k$ represents the detection delay of the $k$-th break-point on arm $i$. Thus, concerning an arm $i$, the total delay up to time $T$ follows by summing $D_i^k$ for each $k \leq \gamma_T$. Let $l_0,l_1,...$ be the length of intervals between detected changes. Hence, in total, we have ($\gamma_T + F_i$) changes and $ \sum_{m=0}^{\gamma_T + F_i} l_m = T$. Assuming $R(T)=O(\log T)$ and by applying the inequality of arithmetic and geometric means (AM-GM inequality) we have,
\begin{align}
R_{tot}(T) &= \sum_{m=0}^{\gamma_T+F_i} R(l_m)+ \sum_{i=1}^{\gamma_T} D_i^k\notag \\
& \leq O(\log\prod_{m=0}^{\gamma_T+F_i}l_m )+ \sum_{i=1}^{\gamma_T} D_i^k \notag \\
& \leq O(\log((\frac{\sum_{m=0}^{\gamma_T+F_i}l_m}{1 + \gamma_T+F_i})^{1+\gamma_T+F_i})+\gamma_T \mathbbm{E}[D_{\varDelta}] \notag\\
& \leq O(1+\gamma_T+\mathbbm{E}[F_i])(\log \frac{T}{1 + \gamma_T+\mathbbm{E}[F_i]})+\gamma_T \mathbbm{E}[D_{\varDelta}]\notag
\end{align}
%
\bibliographystyle{IEEEtran.bst}
\bibliography{references}
\end{document}